\documentclass[conference]{IEEEtran}
\usepackage{cite}
\usepackage{amsmath,amssymb,amsfonts}
\usepackage{algorithmic}
\usepackage{graphicx}
\usepackage{textcomp}
\usepackage{xcolor}

\usepackage[ruled]{algorithm2e}

\SetCommentSty{mycommfont}

\usepackage{subcaption}
\usepackage{multirow}
\usepackage{pifont}

\usepackage{capt-of,etoolbox}
\usepackage{lipsum}

\usepackage{soul}
\usepackage{cite}

\def\BibTeX{{\rm B\kern-.05em{\sc i\kern-.025em b}\kern-.08em
    T\kern-.1667em\lower.7ex\hbox{E}\kern-.125emX}}
\begin{document}

\title{Looking Beyond Corners: Contrastive Learning of Visual Representations for Keypoint Detection and Description Extraction}

\author{
\IEEEauthorblockN{Henrique Siqueira\IEEEauthorrefmark{1}, Patrick Ruhkamp\IEEEauthorrefmark{1}\IEEEauthorrefmark{2}, Ibrahim Halfaoui\IEEEauthorrefmark{1}\IEEEauthorrefmark{2}, Markus Karmann\IEEEauthorrefmark{1,2} and Onay Urfalioglu\IEEEauthorrefmark{1}}
\IEEEauthorblockA{\textit{Munich Research Center, Huawei\IEEEauthorrefmark{1} and Technical University of Munich\IEEEauthorrefmark{2}} \\
Munich, Germany\\
first.second@huawei.com\IEEEauthorrefmark{1} - first.second@tum.de\IEEEauthorrefmark{2}}
}

\maketitle

%------------------------------------------------------------------------------------------------------------------------------------------------------------------------------------------------------------------------
% ----- Abstract
%------------------------------------------------------------------------------------------------------------------------------------------------------------------------------------------------------------------------
\begin{abstract}
Learnable keypoint detectors and descriptors are beginning to outperform classical hand-crafted feature extraction methods. Recent studies on self-supervised learning of visual representations have driven the increasing performance of learnable models based on deep networks. By leveraging traditional data augmentations and homography transformations, these networks learn to detect corners under adverse conditions such as extreme illumination changes. However, their generalization capabilities are limited to corner-like features detected a priori by classical methods or synthetically generated data.

In this paper, we propose the Correspondence Network (CorrNet) that learns to detect repeatable keypoints and extract discriminative descriptions via unsupervised contrastive learning under spatial constraints. Our experiments show that CorrNet is not only able to detect low-level features such as corners, but also high-level features that represent similar objects present in a pair of input images through our proposed joint guided backpropagation of their latent space. Our approach obtains competitive results under viewpoint changes and achieves state-of-the-art performance under illumination changes.
\end{abstract}


%------------------------------------------------------------------------------------------------------------------------------------------------------------------------------------------------------------------------
% ----- Introduction
%------------------------------------------------------------------------------------------------------------------------------------------------------------------------------------------------------------------------
\section{Introduction}
\begin{quotation}
    \begin{flushright}
        \textit{``The AI revolution will not be supervised''}.
        \\- A. Efros~\cite{efros}.
    \end{flushright}
\end{quotation}
\begin{figure}[t]
    \begin{center}
        \includegraphics[width=\columnwidth]{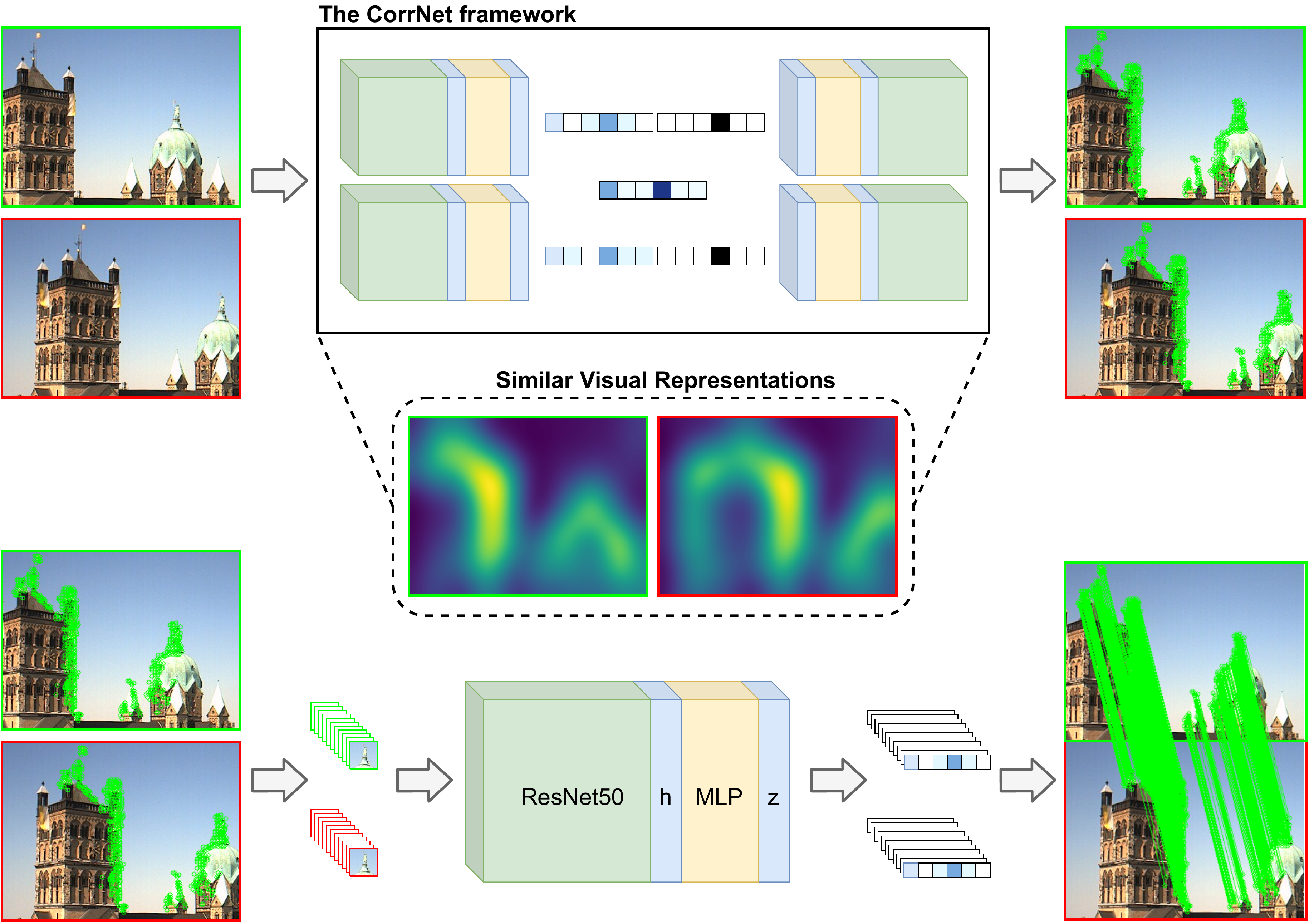}
    \end{center}
    \caption{The CorrNet framework. Illustration of joint keypoint detection and description extraction with the proposed approach. \textit{Top:} CorrNet encodes input images to latent feature vectors. With a novel approach to the guided grad-CAM, CorrNet detects keypoints in common regions between a pair of images in the input space. \textit{Bottom:} The same network can be used for matching local patches centred at the keypoints.}
    \label{fig:teaser}
\end{figure}

Keypoint detection and description extraction in images are important tasks for many real-world applications including pose estimation~\cite{dusmanu2019d2, revaud2019r2d2, 2020_jau_zhu_deepFEPE} and object tracking~\cite{hu2021monocular}. Traditionally, those tasks are performed by extracting hand-crafted features based on prior information such as gradient-based methods~\cite{lowe2004distinctive, bay2006surf}. With the success of deep learning in computer vision, recent approaches that rely on learnable features by training deep neural networks have rapidly gained popularity~\cite{detone2018superpoint, mishchuk2017working}. Most learnable methods, however, are trained under a (semi-)supervised regime and require a huge amount of labelled data from some source~\cite{detone2018superpoint, Barroso-Laguna2019ICCV}.

In fact, supervised learning approaches depend heavily on the amount and quality of labelled training data available to perform well under real-world conditions~\cite{detone2018superpoint, deng2009imagenet}. This limiting factor has encouraged the scientific community to explore alternatives. Self-supervised methods, in particular via contrastive learning (CL), have shown promising results and are potential candidates to enhance generalization performance using a few~\cite{chen2020simple, chen2020big} or no labels at all ~\cite{radford2learning}.

Motivated by the latest achievements in contrastive learning, we present a novel paradigm for detecting repeatable and descriptive keypoints - beyond extracting sole corner and edge detections - by learning visual feature representations from weakly augmented image pairs in a fully unsupervised manner. By training the proposed Correspondence Network (CorrNet) under a contrastive regime with spatial constraints from weakly augmented image pairs, CorrNet neither relies on additional knowledge of the scene and camera poses~\cite{dusmanu2019d2, revaud2019r2d2}, nor strong priors such as homographic viewpoint changes~\cite{detone2018superpoint, Barroso-Laguna2019ICCV}, supervised pre-training ~\cite{detone2018superpoint}, or hand-crafted detection anchors~\cite{Barroso-Laguna2019ICCV} which are typical methods adopted to train modern feature extractors. In our approach, keypoints are detected based on novel guided grad-CAM~\cite{springenberg2015striving, selvaraju2017grad} of CorrNet's latent space. Feature descriptions are extracted by computing representations of local patches surrounding detected keypoints with the identical network.

%------------------------------------------------------------------------------------------------------------------------------------------------------------------------------------------------------------------------
% ----- Related Work
%------------------------------------------------------------------------------------------------------------------------------------------------------------------------------------------------------------------------
\section{Related Work}
Feature extraction (i.e., keypoint detection and description extraction) is a prerequisite for several computer vision tasks. Early approaches ~\cite{harris1988combined, lowe2004distinctive, bay2006surf, nixon2019feature} proposed the use of hand-crafted feature extraction methods for corner and edge detection, followed by description extraction of local information. However, their generalization capabilities are limited and their performance often deteriorates under adverse conditions such as illumination changes~\cite{detone2018superpoint, sarlin20superglue, Barroso-Laguna2019ICCV, nixon2019feature}.

In order to improve generalization performance for real-world applications, classical supervised learning approaches~\cite{rosten2006machine, rosten2008faster, alcantarilla2011fast, rublee2011orb} were applied for feature extraction, enhancing generalization by becoming invariant to various input variations including scale, rotation, and illumination in a sequential pipeline from keypoint detection to description extraction. Sequential approaches for feature extraction are still vastly adopted in computer vision~\cite{simo2015discriminative, tian2017l2, balntas2016learning, mishchuk2017working, luo2018geodesc, luo2019contextdesc}, but at present, the disadvantage of the performance of the latter (description extraction) being tightly dependent on the performance of the former (keypoint detection) limits capabilities and applicability.

Another paradigm for learnable feature extraction is to jointly train a model to detect keypoints and extract descriptions~\cite{yi2016lift, noh2017large, detone2018superpoint, ono2018lf, dusmanu2019d2, revaud2019r2d2}. For instance, in the work of DeTone et al.~\cite{detone2018superpoint}, a multi-head deep neural network (i.e., SuperPoint) is trained under a multitask learning setting where one head minimizes the location of keypoints and the other minimizes a similarity measure of description vectors with respect to the detected keypoints with weak supervision. The labels used to train SuperPoint are synthetically generated through homography transformations in the training images. Their approach achieved state-of-the-art results on the well-established HPatches benchmark~\cite{balntas2017hpatches}, and hence, is defined as a baseline in our experiments.

%------------------------------------------------------------------------------------------------------------------------------------------------------------------------------------------------------------------------
% ----- The Correspondence Network Framework
%------------------------------------------------------------------------------------------------------------------------------------------------------------------------------------------------------------------------
\section{The Correspondence Network Framework}
%------------------------------------------------------------------------
%--- Introduction to CorrNet and section's structure
%------------------------------------------------------------------------
The Correspondence Network (CorrNet) is an unsupervised visual representation learning approach for joint keypoint detection and description extraction. Unlike the majority of contrastive learning-based methods which use self-supervision as a precursory mechanism to improve performance on a subsequent task (e.g., classification), CorrNet leverages contrastive learning as the main training objective. We hypothesize that a deep neural network trained by contrasting similar pairs of images is able to learn \textit{repeatable} and \textit{discriminative} visual features. These features describe the similarity between two images and recurrent patterns inherent in the training set. Consequently, they could be used as keypoints and their latent representations as descriptions.

To detect repeatable and discriminative features in the input space, we propose an alternative approach to the guided grad-CAM algorithm~\cite{selvaraju2017grad} that is best tailored to approaches for metric learning such as CorrNet. In this section, we describe the CorrNet training procedure using contrastive learning under spatial constraints. Then, we conclude the section by describing the full pipeline (shown in Figure~\ref{fig:cornet_framework}) using first a CorrNet model for keypoint detection and global description, and then a fine-tuned one for discriminative local descriptions of the previously detected keypoints.

%------------------------------------------------------------------------
%--- How to train cornet
%------------------------------------------------------------------------
\subsection{Contrastive Learning with Spatial Constraints}
\begin{figure}
    \begin{center}
        \includegraphics[width=\columnwidth]{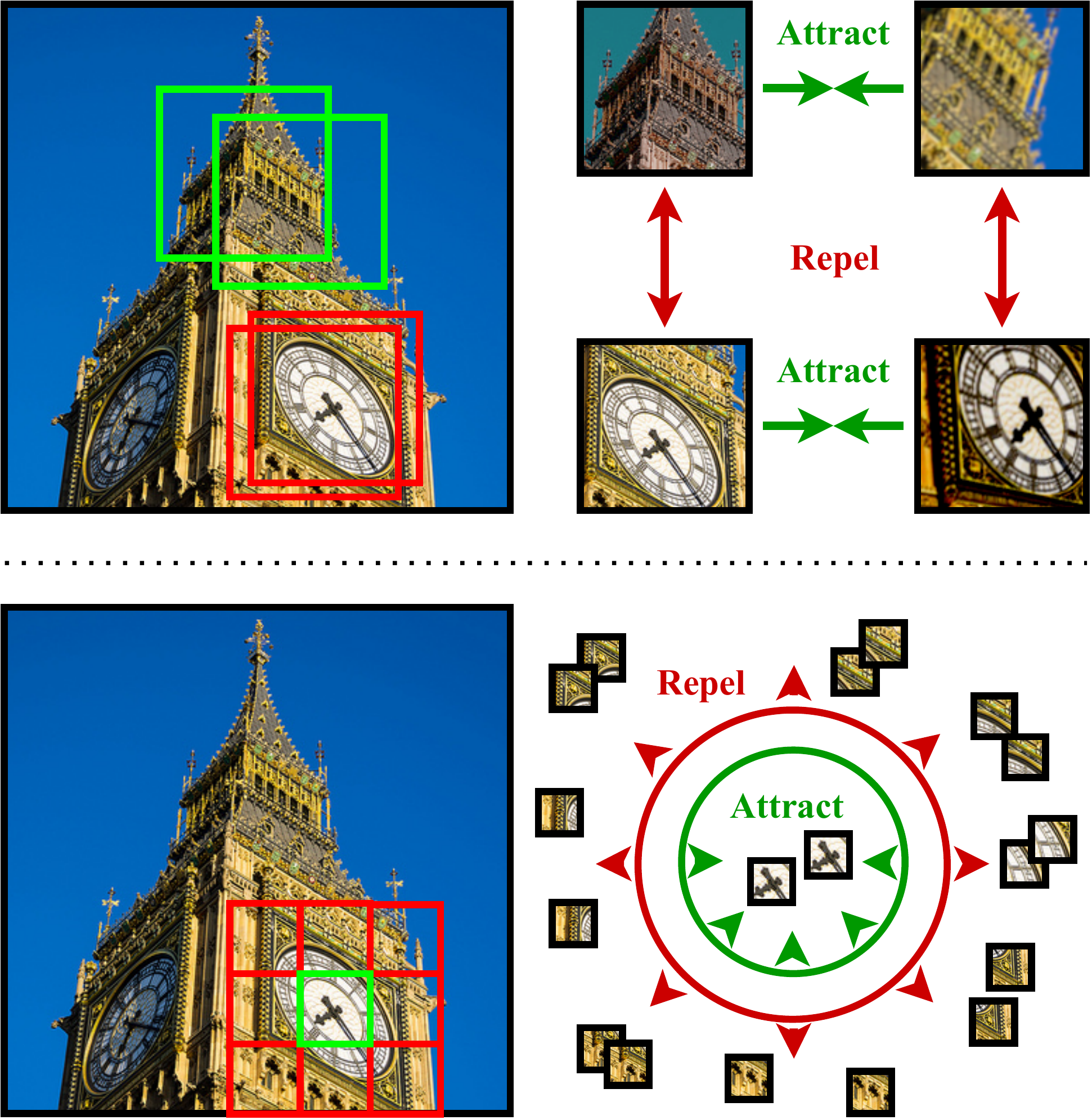}
    \end{center}
    \caption{The CorrNet contrastive learning paradigm. 
            \textit{Top:} Large crops are extracted from an image to build positive and negative examples to learn visual feature representations with CorrNet for keypoint detection.
            \textit{Bottom:} A multitude of small negative crops are sampled around a positive small crop to train a feature descriptor with CorrNet.}
    \label{fig:training}
\end{figure}

Motivated by the latest results on self-supervised learning~\cite{chen2020big}, the CorrNet framework consists of a siamese convolutional neural network trained via contrastive learning. The objective is to automatically learn visual similarities and differences between image pairs by exploiting ideas from SimCLR~\cite{chen2020simple}. Unlike previous methods~\cite{schroff2015facenet}, SimCLR does not require accurate sample selection. However, without proper data augmentation, the network tends to fail to learn useful visual representations as demonstrated by Chen et al.~\cite{chen2020simple}. As a result, the performance of the network in subsequent tasks is negatively affected. In order to be able to detect highly-repeatable keypoints and extract discriminative local descriptions, the network should learn representations of the most salient features and their spatial information. To do so, we propose the application of spatial constraints in contrastive learning. The spatial constraints are modelled as soft constraints to foster the model to learn common and salient visual features between image pairs~\cite{gori2017machine}.

\textbf{Fostering highly repeatable keypoints.} Aiming to preserve spatial relationships between features from image pairs, a positive training pair is defined as two random crops with overlapping regions, as shown by the green crops in Figure~\ref{fig:training} (top illustration). To foster CorrNet to learn efficient representations for feature extraction and avoid learning trivial solutions~\cite{zhang2016colorful, chen2020simple}, we increase the difficulty of the optimisation objective by taking another pair from the same image without overlapping regions with respect to the first pair (red crops). These spatial constraints are the main differences between CorrNet and SimCLR. In the latter, the green and red crops are considered positive examples since both represent features of the same object (i.e., Big Ben), whereas in the former a pair of the green and red crops are defined as a negative pair because there are no repeatable keypoints between them. The other differences are photometric and weak geometric transformations applied to each image to make CorrNet invariant to input changes caused by illumination or perspective variations.

\textbf{Fostering highly discriminative descriptions.} As training progresses, the network learns gradually to become invariant to strong input changes due to the data augmentations. Assuming that two keypoints are close to each other in a reference image with common neighbouring pixels, a network trained to become invariant to this sort of situation would generate ambiguous description vectors. This is detrimental to the task of finding corresponding keypoints in the target image. Thus, we propose to train an additional set of weights for CorrNet under neighbouring spatial constraints to enhance discriminative descriptions as illustrated in Figure~\ref{fig:training} (bottom). Under the neighbouring constraints, a positive pair is defined as crops of local regions centred at detected keypoints (the green crop) and the neighbouring local crops in red are considered to be the negative samples with respect to green crops.

CorrNet is trained to minimize the normalized temperature-scaled cross-entropy loss function (NT-Xent)~\cite{chen2020simple}, defined for a single positive pair as follows:
\begin{equation}
    \begin{aligned}
        l_{i,j} = -\log \dfrac{\exp(\mathrm{sim}(z_i,z_j)/\tau)}{\sum^{2N} \limits_{k=1} {1}_{[k \ne i]}  \exp(\mathrm{sim}(z_i,z_k)/\tau)} 
    \end{aligned}
\end{equation}
A positive pair of similar transformed input images $(x_i, x_j)$ are presented to the network to compute their respective description vectors $z_i$ and $z_j$. The exponential cosine similarity between the description vectors of the positive pair $(z_i,z_j)$ is divided by the summation of the exponential cosine similarities from negative pairs $(z_i,z_k)$ where $x_i \ne x_k$. The temperature parameter $\tau$ is utilized for normalization.

%------------------------------------------------------------------------
%--- Keypoint Detection and Description Extraction
%------------------------------------------------------------------------
\subsection{Joint Guided Gradient Backpropagation}
\begin{figure*}[!ht]
    \centering
    \includegraphics[width=\textwidth]{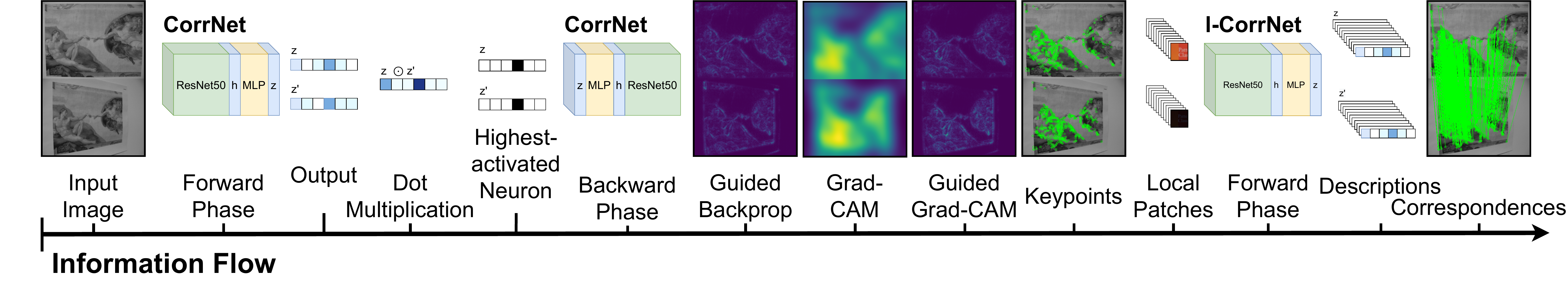}
    \caption{Inference phase for CorrNet. CorrNet encodes visual feature representations to latent vectors. The most activated and common neurons in the vectors (with the same neuron index) are backpropagated to the input space to generate keypoints located in regions with similar visual content. Moreover, local patches around the keypoints are fed to the same architecture for description extraction and matching. Please note the overlap of keypoint activation regions on the top and bottom images in the \textit{GradCam} stage despite the different scene perspectives.}
    \label{fig:cornet_framework}
\end{figure*}
Figure~\ref{fig:cornet_framework} depicts the complete inference or processing pipeline for joint keypoint detection and description extraction with the CorrNet framework. The reference and target images ($x$ and $x'$) are introduced to the model to compute their description vectors ($z$ and $z'$). In the case of joint input including a reference and a target image, keypoints are detected by applying a modified version of the guided grad-CAM algorithm~\cite{selvaraju2017grad} on the CorrNet's latent units.

The original algorithm combines the guided backpropagation algorithm for fine-grained salient map generation, and the grad-CAM algorithm to reduce the noise of the former and highlight regions with more semantic content. Guided grad-CAM~\cite{selvaraju2017grad} has been utilized to indicate units in the input space responsible for the high activation of a target neuron in the output layer. In their algorithm, activations are backpropagated with respect to the same image.

In our version of guided grad-CAM, we perform operations in the latent space to correlate the visual representations of the reference and target input images before the backpropagation of the activations. More specifically, we multiply the latent vector of the reference image to the last feature maps of the target image and vice-versa as a modulation factor for our algorithm to focus attention on common visual representations.

\begin{figure}[!ht]
    \begin{center}
        \includegraphics[width=\columnwidth]{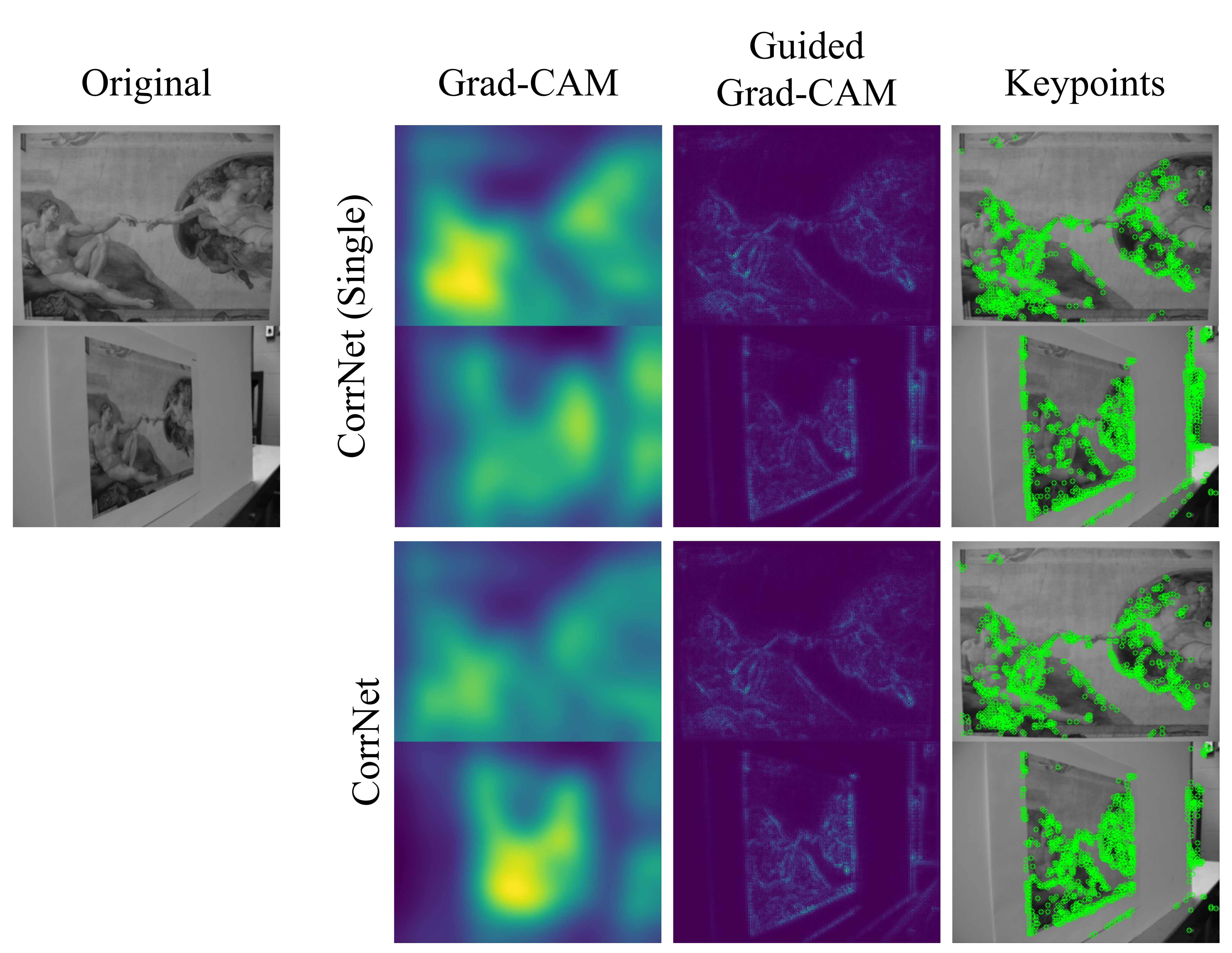}
    \end{center}
    \caption{A comparison of the standard guided backprop (CorrNet (Single), 55.7\% of repeatable keypoints) and our novel approach (CorrNet, 71.0\% of repeatable keypoints).}
    \label{fig:cam}
\end{figure}

Moreover, we multiply the latent vectors $z$ and $z'$ to identify the neuron that retains information about similar visual features. We define the neuron with the highest value as the target for visualization. We hypothesize that the high activation of this neuron is due to the presence of common visual features in the input space in related images $x$ and $x'$. The most activated neuron between of the latent vector is backpropagated to the input space to generate keypoints on the reference and target images in regions with similar content. A qualitative impression of our algorithm is illustrated in Figure~\ref{fig:cam}. Note that our novel approach can robustly filter out many noisy keypoints in the non-overlapping regions in the background with non-similar image content. The keypoints are mostly detected in these regions with similar content.

Afterwards, we apply non-maximum suppression on the gradient images in the guided grad-CAM stage in Figure~\ref{fig:cornet_framework} to reduce the number of keypoints in a small region and select the top-k units with the largest gradients. Our assumption is that these units in the input space have the highest likelihood to be found on both the reference and target images.

Finally, the last part of the correspondence network framework extracts local patches centred at the detected keypoints. Only these patches are used as input to l-CorrNet to compute its descriptions. The cosine similarity between the local description vectors from l-CorrNet can be used for matching.

%------------------------------------------------------------------------------------------------------------------------------------------------------------------------------------------------------------------------
% ----- Experiments
%------------------------------------------------------------------------------------------------------------------------------------------------------------------------------------------------------------------------
\section{Experiments}
In this section, we systematically evaluate CorrNet as a joint detector and descriptor under different experimental conditions. We follow the experimental protocol defined by DeTone et al.~\cite{detone2018superpoint}  - thus adopting the resolution of 240x320 - which includes the use of the MS-COCO~\cite{lin2014microsoft} as a training set and HPatches~\cite{balntas2017hpatches} as a test set. In the following, we describe the evaluation metrics and the implementation details. Afterwards, we present and discuss quantitative and qualitative results, as well as extensive ablation studies on keypoint detection. 

%------------------------------------------------------------------------
%--- Evaluation metrics
%------------------------------------------------------------------------
\subsection{Evaluation Metrics}
Ideally, keypoints should be repeatable under various scene and image conditions, such as image noise, illumination and perspective changes. A widely used criterion is repeatability as introduced by Mikolajczyk and Schmid~\cite{mikolajczyk2004scale}. It represents the ratio between the number of repeatable keypoints over the total number of detections defined as follows:

\begin{equation}
\begin{aligned}
\textrm{REP} = \dfrac{1}{N_1 + N_2} \left(\sum\limits_{i} \gamma(x_i) + \sum\limits_{j} \gamma(x_j)\right).
\end{aligned}
\end{equation}

As a second metric, the localization error is used to evaluate the accuracy of the position of detected keypoints as follows: 
\begin{equation}
\begin{aligned}
\textrm{LE} = \dfrac{1}{N} \sum\limits_{i:\gamma(x_i)} \min\limits_{j \in \{1,\dots, K\}} || x_i - \hat{x}_i||,
\end{aligned}
\end{equation}
where correctness $\gamma(x_i)$ is defined with respect to the threshold $\epsilon$ 
depicting the maximum correct distance between two points, having $N_1$ points in the first image and $N_2$ points in the second image. We define correctness~\cite{detone2018superpoint} as:

\begin{equation}
\begin{aligned}
\gamma(x_i) =  \left(\min\limits_{j \in \{1,\dots, N_2\}} || x_i - \hat{x}_i||\right) \leq \epsilon.
\end{aligned}
\end{equation}

The third metric is based on the estimated homography $\mathrm{\hat{H}}$ against the ground-truth homography $\mathrm{H}$. We evaluate how accurately the homography transforms the four corners of one image onto the other. If the reference image has corners $c_1, c_2, c_3, c_4$ respectively, the ground-truth $\mathrm{H}$ homography is applied to recover the ground-truth corners in the target image $c'_1, c'_2, c'_3, c'_4$ and the estimated homography $\mathrm{\hat{H}}$ to get $\hat{c}'_1, \hat{c}'_2, \hat{c}'_3, \hat{c}'_4$. The threshold $\epsilon$ is used to denote a correct homography and resulting scores should range between 0 and 1 where higher is always better.

\begin{equation}
\begin{aligned}
\textrm{CorrH} = \dfrac{1}{N} \sum\limits_{i=1}^{N} \left(\left( \sum\limits_{j=1}^{4} || c_{i,j} - \hat{c}'_{i,j}|| \right)\leq \epsilon  \right)
\end{aligned}
\end{equation}

Finally, we rely on the publicly available evaluation scripts from Jau~\cite{2020_jau_zhu_deepFEPE}\footnote{https://github.com/eric-yyjau/pytorch-superpoint} and Sarlin~\cite{sarlin20superglue}\footnote{https://github.com/rpautrat/SuperPoint}. In these scripts, keypoints detected in a reference image that would correspond to a location outside the target image (and vice-versa) are filtered out using the ground-truth homography provided in the HPatches dataset. Although this practice has been previously used in the literature~\cite{detone2018superpoint, Barroso-Laguna2019ICCV}, the use of the ground-truth homography would be unrealistic since it is not available in real-world applications. Therefore, this filtering method is omitted in our experiments in order to present results that better reflect the performance of the approaches in real-world applications.

%------------------------------------------------------------------------
%--- CorrNet's Implementation Details
%------------------------------------------------------------------------
\subsection{Implementation Details}
Our architecture consists of a siamese convolutional neural network. The CorrNet encoder, as illustrated in Figure~\ref{fig:cornet_framework}, is a ResNet50 with no max-pooling layers and pre-trained weights of SimCLR\cite{chen2020simple}\footnote{https://github.com/google-research/simclr}. The projection head, on the other hand, was not made available. In our approach, it comprises a multilayer perceptron of three layers with 2048, 512, and $d$ neurons, respectively. The parameter $d$ denotes the size of the description vector. The Adam algorithm~\cite{Adam} was used to optimize CorrNet. After extensive exploratory experiments, we fixed the learning rate to $1e^{-3}$, weight decay to $1e^{-6}$, batch size of 200 and the temperature parameter of the NT-Xent loss to 0.5. The code is publicly available in our repository\footnote{https://github.com/siqueira-hc/CorrNet}.

%------------------------------------------------------------------------
%--- Keypoints: comparison with the literature
%------------------------------------------------------------------------
\subsection{Evaluation of the Keypoints}
%------------------------------------------------------------------------
%--- Discussion
%------------------------------------------------------------------------
Table~\ref{tab:sota_hpatches} presents the performance of our approach, CorrNet, in comparison with state-of-the-art methods for keypoint detection. To conduct a fair evaluation for all of the methods, they are strictly evaluated under the same condition. In summary, the input image size is 240x320, and non-maximum suppression (NMS) of 3x3 is applied to increase the distribution of the detected key points in the image. In addition, no filtering method using ground truth is utilized, and the top 1000 keypoints are detected in each image. Moreover, only original implementations from their official repositories are used such as SuperPoint\footnote{https://github.com/magicleap/SuperPointPretrainedNetwork}, SuperPoint (Gauss)\footnote{https://github.com/eric-yyjau/pytorch-superpoint}, and Key.Net\footnote{https://github.com/axelBarroso/Key.Net}. For the classical methods, we use the OpenCV implementations.

CorrNet achieves state-of-the-art results for illumination changes on both repeatability and localization error, demonstrating its robustness against adverse environmental conditions, and competitive results on viewpoint changes. Surprisingly, consistent with recent findings from Balntas and Lenc et al.\cite{balntas2017hpatches}, classical hand-crafted methods can still play an important role in modern applications. Moreover, CorrNet's superior results when compared against SimCLR (pre-trained network from Chen et al.~\cite{chen2020simple} using our guided grad-CAM that correlates two input images by exchanging the grad-CAM flow between images) and ELF~\cite{benbihi2019elf} (a gradient-based method for keypoint detection using pre-trained networks) show that the proposed training method via contrastive learning under spatial constraints provides a significant performance improvement and plays an important role in detecting highly repeatable keypoints.

\begin{table}[h]
\caption{Comparison with state of the art. Repetability (rep.) and localization error (loc. error) of keypoints on HPatches for illumination and viewpoint changes. \textit{se}: semi-supervised,  \textit{us}: unsupervised, and \textit{hc}: hand-crafted.}
\begin{center}
\resizebox{\linewidth}{!}{\begin{tabular}{||l|l|c|c|c|c||}
\hline
\multirow{2}{*}{Approach} & \multirow{2}{*}{Nature} & \multicolumn{2}{c|}{Illumination} & \multicolumn{2}{c||}{Viewpoint} \\ \cline{3-6}
& & Rep. $\uparrow$ & Loc. Error          & Rep.              & Loc. Error        \\ \hline \hline
CorrNet (ours)                          & us            & $\textbf{77.2\%}$ & $\textbf{0.86}$   & $63.3\%$          & $1.38$            \\ \hline
CorrNet (ours, single)                  & us            & $76.3\%$          & $0.87$            & $59.9\%$          & $1.39$            \\ \hline
Harris~\cite{harris1988combined}        & hc            & $74.0\%$          & $1.03$            & $\textbf{66.6\%}$ & $1.18$            \\ \hline
Shi-Tomasi~\cite{shi1994good}           & hc            & $73.2\%$          & $1.13$            & $65.2\%$          & $1.25$            \\ \hline
FAST~\cite{rosten2008faster}            & hc            & $72.9\%$          & $1.02$            & $65.7\%$          & $1.21$            \\ \hline
SuperPoint (Gauss)~\cite{2020_jau_zhu_deepFEPE} & se    & $72.4\%$          & $1.20$            & $56.1\%$          & $1.40$            \\ \hline
SuperPoint~\cite{detone2018superpoint}  & se            & $71.5\%$          & $1.22$            & $55.8\%$          & $1.45$            \\ \hline
ELF~\cite{benbihi2019elf}               & -             & $71.3\%$          & $1.45$            & $60.0\%$          & $1.59$            \\ \hline
Key.Net~\cite{Barroso-Laguna2019ICCV}   & se            & $70.6\%$          & $0.90$            & $59.0\%$          & $\textbf{1.06}$   \\ \hline
ASLFeat~\cite{luo2020aslfeat}           & se            & $67.9\%$          & $1.12$            & $55.0\%$          & $\textbf{1.20}$   \\ \hline
R2D2~\cite{revaud2019r2d2}              & us            & $67.2\%$          & $1.32$            & $46.3\%$          & $1.43$            \\ \hline
SimCLR~\cite{chen2020simple}            & us            & $58.2\%$          & $1.09$            & $38.4\%$          & $1.55$            \\ \hline
Random                                  & hc            & $32.5\%$          & $1.94$            & $26.2\%$          & $1.90$            \\ \hline
\end{tabular}}
\end{center}
\label{tab:sota_hpatches}
\end{table}

\begin{figure*}[!ht]
    \centering
    \includegraphics[width=\textwidth]{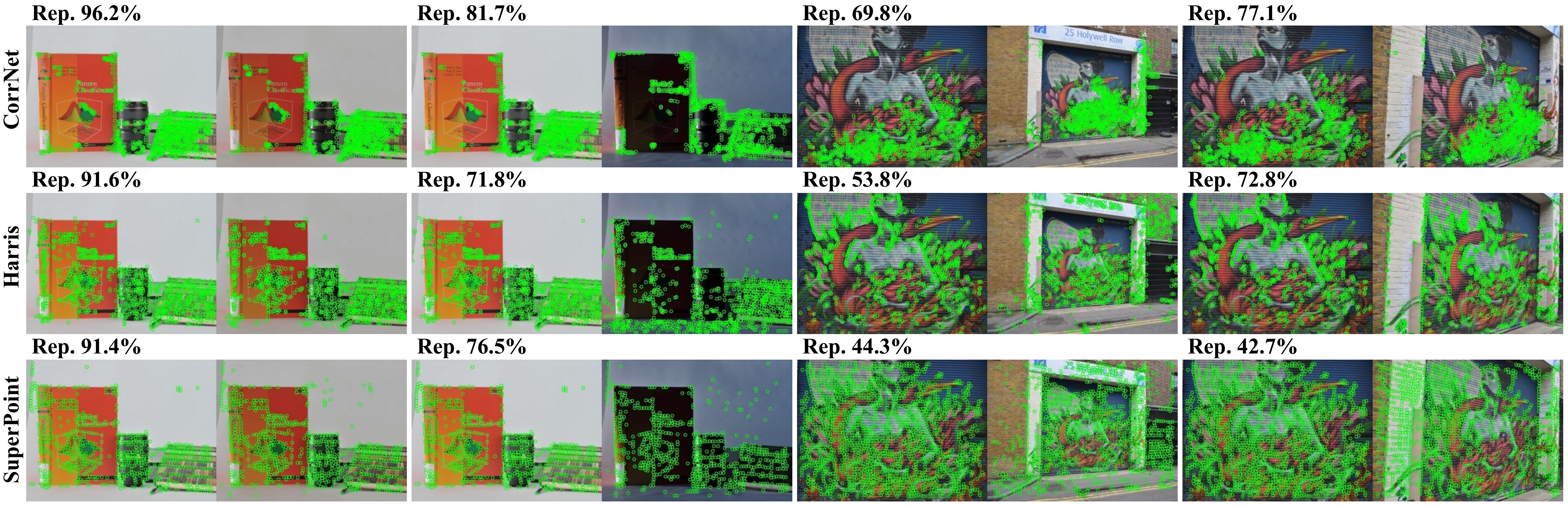}
    \caption{Qualitative results of keypoint detection using CorrNet, Harris~\cite{harris1988combined} and SuperPoint~\cite{detone2018superpoint}. Repeatability scores are shown on top of each pair of images. On the left results under illumination changes, and on the right results under viewpoint changes.}
    \label{fig:qua_key}
\end{figure*}

\begin{figure*}[!ht]
    \centering
    \includegraphics[width=\textwidth]{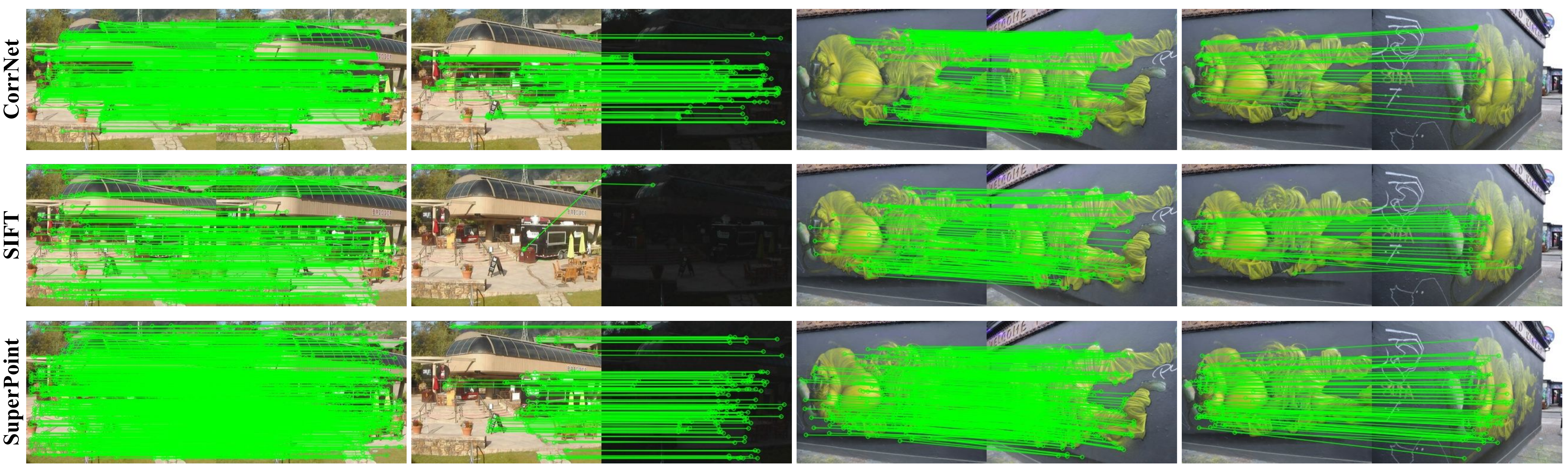}
    \caption{Qualitative results of keypoint matching using keypoints and descriptions from CorrNet, SIFT\cite{lowe2004distinctive} and SuperPoint~\cite{detone2018superpoint}. On the left results under illumination changes, and on the right results under viewpoint changes.}
    \label{fig:qua_homo}
\end{figure*}

Figure~\ref{fig:qua_key} presents qualitative results on repeatability from CorrNet, Harris, and SuperPoint detections. While CorrNet is able to detect repeatable and stable keypoints under different illumination changes, Harris and SuperPoint suffer from the challenging illumination condition in the target image where keypoints are detected in dark regions due to the false-positive corner and edge detection caused by shades. Under different viewpoints, CorrNet achieves higher repeatability scores on these examples by successfully focusing the detection on similar objects present in the reference and target images, whereas the other methods blindly detect keypoints on the entire images.

Finally, it is important to notice that although CorrNet is trained in a fully unsupervised manner with weakly augmented image pairs, our generic training scheme can yield comparable results on viewpoint changes. In fact, our approach even outperforms semi- and self-supervised methods such as SuperPoint~\cite{detone2018superpoint, 2020_jau_zhu_deepFEPE}, ELF~\cite{benbihi2019elf}, ASLFeat~\cite{luo2020aslfeat}, and R2D2~\cite{revaud2019r2d2}. We believe further research including perspective viewpoint changes, and exploiting the temporal domain of sequences would lead to further improvements.

%------------------------------------------------------------------------
%--- Descriptions: comparison with the literature
%------------------------------------------------------------------------
\subsection{Evaluation of the Descriptions}
%------------------------------------------------------------------------
%--- Discussion
%------------------------------------------------------------------------
In this experiment, we evaluate l-CorrNet as a description extraction method and compare our results with the state-of-the-art. Table~\ref{tab:homography_estimation} summarizes the success rate of homography estimation under different thresholds: 1, 3, and 5. Under illumination changes, l-CorrNet (single) achieves the most accurate homography estimation with a threshold of 1 by a large margin, and similar results compared to HardNet under a threshold of 3. With a threshold of 5 l-CorrNet achieves comparable results to HardNet. For homography estimation from different viewpoints, SIFT shows to be the most reliable and accurate method. We observe l-CorrNet (single) achieves better results compared to l-CorrNet for homography estimation. Our hypothesis is that keypoints that are essential for the homography computation may be located within regions that are less relevant in terms of similar visual features and are therefore not entirely recovered. Additionally, large homographic changes may lead to only a small overlap between image pairs, which would foster keypoint clusters with our novel approach, making it more challenging for homography computation.

\begin{table}[h]
\caption{Homography estimation errors under illumination and viewpoint changes and different correctness criteria. \textit{se}: semi-supervised, \textit{us}: unsupervised, and \textit{hc}: hand-crafted.}
\begin{center}
\resizebox{\linewidth}{!}{\begin{tabular}{||l|l|c|c|c|c|c|c||}
\hline
\multirow{2}{*}{Approach} & \multirow{2}{*}{Nature}  & \multicolumn{3}{c|}{Illumination} & \multicolumn{3}{c||}{Viewpoint} \\ \cline{3-8}
                                            &     & $\epsilon = 1\uparrow$ & $\epsilon = 3$  & $\epsilon = 5$  & $\epsilon = 1$  & $\epsilon = 3$  & $\epsilon = 5$  \\ \hline \hline
CorrNet (ours, single)  & us  & $\textbf{84.6\%}$               & $\underline{94.7\%}$        & $\underline{97.2\%}$        & $11.2\%$        & $37.3\%$        & $52.2\%$    \\ \hline
CorrNet (ours)   & us          & $71.9\%$               & $92.6\%$        & $96.8\%$        & $5.1\%$        & $33.2\%$        & $46.1\%$        \\ \hline

KeyNet~\cite{Barroso-Laguna2019ICCV} + HardNet~\cite{mishchuk2017working} & se  & $70.6\%$               & $\textbf{96.8\%}$        & $\textbf{98.9\%}$        & $\underline{43.1\%}$        & $\underline{76.3\%}$        & $\underline{83.7\%}$\\ \hline
SIFT~\cite{lowe2004distinctive}   & hc             & $69.1\%$               & $88.4\%$        & $89.8\%$        & $\textbf{59.0\%}$        & $\textbf{79.3\%}$        & $\textbf{85.8\%}$        \\ \hline
SuperPoint~\cite{detone2018superpoint}  & se        & $47.4\%$               & $88.4\%$        & $96.8\%$        & $20.0\%$        & $60.3\%$        & $77.6\%$        \\ \hline
SuperPoint (Gauss)~\cite{2020_jau_zhu_deepFEPE} & se  & $52.6\%$               & $87.0\%$        & $95.4\%$        & $8.5\%$         & $45.8\%$        & $62.0\%$        \\ \hline
ORB~\cite{rublee2011orb}    & hc                     & $36.8\%$               & $64.6\%$        & $72.3\%$        & $11.5\%$        & $50.8\%$        & $60.3\%$        \\ \hline
\end{tabular}}
\end{center}
\label{tab:homography_estimation}
\end{table}

From a qualitative analysis of CorrNet's matching presented in Figure~\ref{fig:qua_homo}, we show CorrNet's robustness against extreme illumination changes. In this second column, SIFT fails to match enough keypoints for accurate homography estimation. SuperPoint achieves similar qualitative results as CorrNet. Nonetheless, keypoints detected by CorrNet are more distributed over the images which often leads to more accurate homography estimation. Under viewpoint changes, the distribution of keypoints over the images may have affected homography estimation using CorrNet when compared against SIFT and SuperPoint (Figure~\ref{fig:qua_homo}, right).

%------------------------------------------------------------------------
%--- Ablation Studies
%------------------------------------------------------------------------
\subsection{Ablation Studies}
Aiming to identify the major factors with a direct impact on CorrNet's performance as a joint keypoint detector and description extractor, we carefully investigate different parameters in the CorrNet framework. The research questions to be studied in this section are the following: (1) what is the impact of the description size on keypoint repeatability? (2) which latent vector, $h$ or $z$, better encodes the visual similarities between two input images, reference and target? (3) What is the impact of different levels of spatial constraints on the repeatability of keypoints? (4) How can the likelihood of the detection of repeatable keypoints be increased? And finally, (5) what is the best set of data augmentation methods for keypoint detection?

\begin{figure}
    \begin{center}
        \includegraphics[width=0.9\columnwidth]{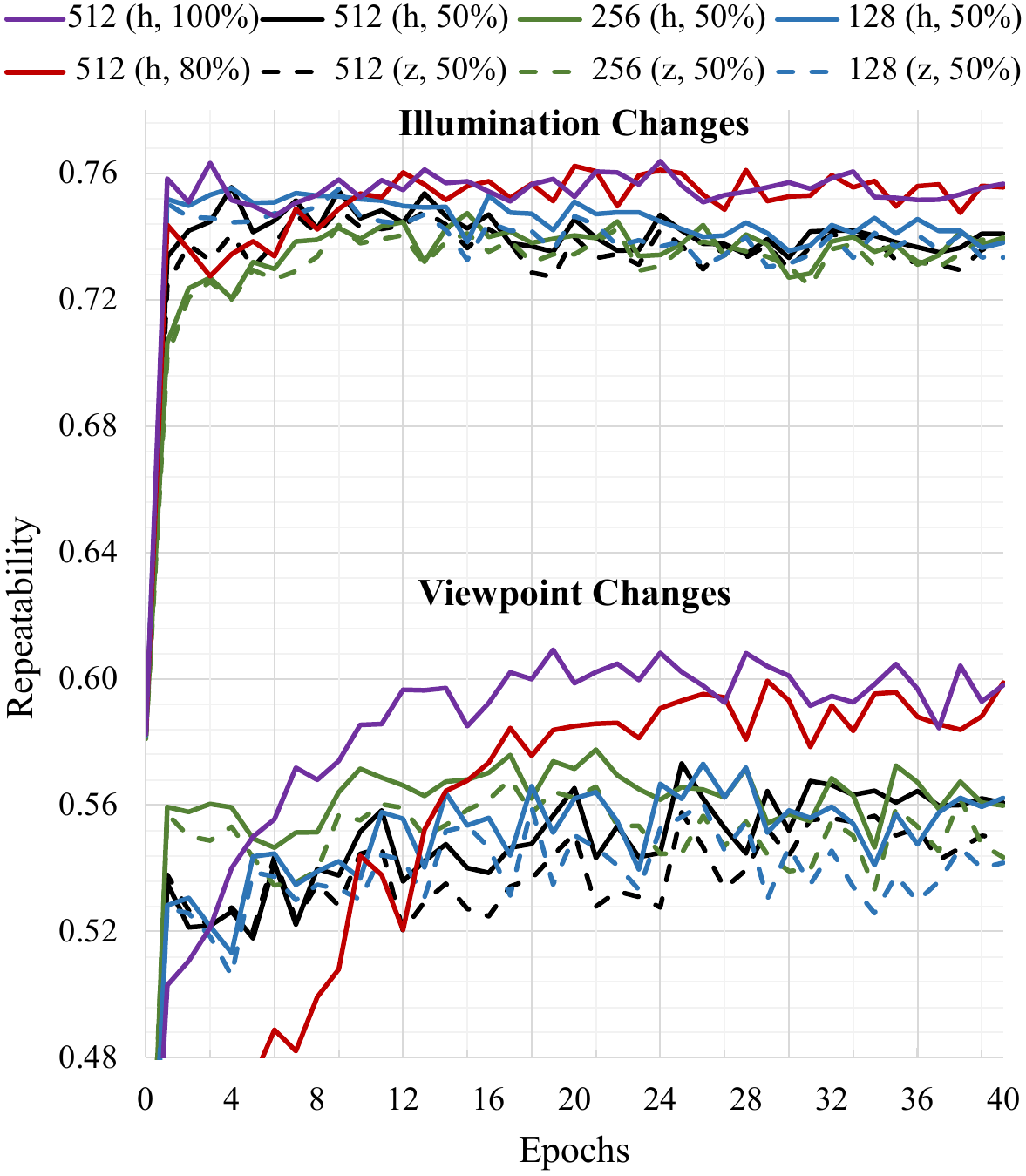}
    \end{center}
    \caption{Repeatability on HPatches over training epochs on MS-COCO for CorrNets with different description size, different spatial constraints ($\%$) and target latent space ($h$ and $z$).}
    \label{fig:description}
\end{figure}

\textbf{Description size and the most informative latent space.} Figure~\ref{fig:description} depicts the training dynamics of CorrNet with respect to different description sizes (128, 256, and 512 units) by showing their repeatability over epochs. Solid lines show the repeatability of keypoints by applying the guided backpropagation with respect to the latent vector $h$, whereas dashed lines with respect to the latent vector $z$. Our results are consistent with the work of Chen et al.~\cite{chen2020simple} which argues that the latent vector $h$ provides more useful visual representations of the data. Indeed, the repeatability scores are consistently higher for all of the CorrNets at different epochs when the guided backpropagation algorithm is applied from $h$. Note CorrNet's low repeatability scores for both illumination ($58.2\%$) and viewpoint ($38.4\%$) in the beginning changes drastically during the first few epochs. These results suggest that the pre-trained SimCLR, while performing highly accurate on ImageNet trained without spatial constraints, is not able to detect a high number of repeatable keypoints when compared to a mature CorrNet trained under the proposed spatial constraints. Regarding the description size, all of the CorrNets show similar behaviours. They achieve high repeatability scores after a few epochs and the difference in performance decreases over time.

\textbf{Contrastive learning under spatial constraints.} Figure~\ref{fig:description} also shows repeatability over training epochs for different CorrNets trained under different spatial constraints on the illumination and viewpoint sets of the HPatches dataset. The only variable factor in this experiment is the proportion of the overlapping region between image pairs indicated by its overlapping percentage. The higher the percentage, the larger the overlapping region between a pair of the positive images. For instance, the training batch for CorrNet (512, 100\%) is always composed of positive pairs with 100\% of overlap. For illumination changes, the repeatability scores are similar but the highest scores are achieved by CorrNet trained under stronger spatial constraints, namely CorrNet (512, 80\%) and CorrNet (512, 100\%). The difference in performance becomes more evident under different viewpoint changes where correspondent keypoints are placed in different coordinates in the reference and target images. Therefore, the spatial information of similar features is crucial to achieving high repeatability scores. CorrNet (512, 100\%) trained under the strongest spatial constraints achieves the best performance on the viewpoint experiments. These results support our initial hypothesis, showing that contrastive learning under spatial constraints plays an important role in the detection of repeatable keypoints.

\begin{table}[]
\caption{Comparison of different CorrNets with the best performance on the validation set of MS-COCO based on the NT-Xent contrastive loss. Repeatability (rep.) and localization error (loc. error) of detected keypoints on HPatches for illumination and viewpoint changes.}
\begin{center}
\resizebox{\linewidth}{!}{\begin{tabular}{||l|c|c|c|c||}
\hline
\multirow{2}{*}{Approach $\downarrow$} & \multicolumn{2}{c|}{Illumination} & \multicolumn{2}{c||}{Viewpoint} \\ \cline{2-5}
                                                & Rep.            & Loc. Error  & Rep.      & Loc. Error    \\ \hline \hline
CorrNet-128                                     & $75.4\%$        & $0.94$      & $59.7\%$  & $1.39$        \\ \hline
CorrNet-128 (single)                            & $73.8\%$        & $0.95$      & $56.8\%$  & $1.40$        \\ \hline
CorrNet-256                                     & $75.5\%$        & $0.90$      & $61.1\%$  & $1.38$        \\ \hline
CorrNet-256 (single)                            & $74.5\%$        & $0.91$      & $58.2\%$  & $1.39$        \\ \hline
CorrNet-512                                     & $74.5\%$        & $0.95$      & $59.4\%$  & $1.38$        \\ \hline
CorrNet-512 (single)                            & $73.6\%$        & $0.96$      & $56.8\%$  & $1.39$        \\ \hline
CorrNet-512 (80\%)                              & $76.8\%$        & $0.89$      & $61.2\%$  & $1.40$        \\ \hline
CorrNet-512 (single, 80\%)                      & $75.7\%$        & $0.90$      & $58.5\%$  & $1.41$        \\ \hline
CorrNet-512 (100\%)                    & $\textbf{77.2\%}$        & $\textbf{0.86}$      & $\textbf{63.3\%}$  & $\textbf{1.38}$        \\ \hline
CorrNet-512 (single, 100\%)                     & $76.3\%$        & $0.87$      & $59.9\%$  & $1.39$        \\ \hline
\end{tabular}}
\end{center}
\label{tab:summary_corrnet}
\end{table}

\textbf{Correlating reference and target images.} Table~\ref{tab:summary_corrnet} shows a systematic comparison between the keypoint detection from CorrNet when using a single image and CorrNet when correlating visual representations of the input features that represent the similarity between the reference and target image. The former approach, indicated in the table by the keyword single, performs the guided backpropagation of the most activated neuron in the $h$ vector, whereas the latter approach is the proposed alternative to the guided grad-cam for CorrNet and metric learning. All of the CorrNets that used our method to correlate reference and target image demonstrated a better performance than the traditional guided backpropagation algorithm. This is evidence that our method increases the likelihood of detection of repeatable keypoints in regions that present similar visual features. In fact, Figure~\ref{fig:cam} presents a qualitative result where regions from the reference image (top) that contains visual features presented in the target image (bottom) are highly activated according to the salience maps. Although keypoints are still detected in regions that are not presented in both, the reference and target image, our method decreased the number of keypoints detected in such regions (see the concentration of keypoints on the right of the image), resulting in an improvement of $15.3\%$ in repeatability. Nevertheless, some less relevant structures are still detected due to sub-optimal representations learned from the dataset, which could be further improved with future research on representation learning.

\textbf{Ablation study on data augmentation techniques.} We evaluate the influence of different data augmentation strategies on detection performance quantitatively. As before, we train on MS-COCO and test on HPatches. We experiment with colour augmentations only, and colour augmentations plus random homography transformations (see Table~\ref{tab:abl_augmentations}.) The random homography transformation is implemented in a manner such that it does not yield out-of-image pixels after being applied to the input image. We perform zooming, which is coupled with the degree of transformation of the homography matrix to avoid artefacts.

\begin{table}[]
\caption{Ablation study on different data augmentation strategies under  illumination and viewpoint changes. Augmentations: \textit{C}: color; \textit{C+H}: color and homography.}
\begin{center}
\resizebox{\linewidth}{!}{\begin{tabular}{||l|c|c|c|c|c||}
\hline
\multirow{2}{*}{Approach} & \multirow{2}{*}{Aug.} & \multicolumn{2}{c|}{Illumination} & \multicolumn{2}{c||}{Viewpoint} \\ \cline{3-6}
                                             &   & Rep.            & Loc. Error  & Rep.      & Loc. Error    \\ \hline \hline
CorrNet-256 (single) & C                                   & $78.8\%$        & $0.75$      & $56.8\%$  & $1.29$        \\ \hline
CorrNet-256 &C                              & $80.8\%$        & $0.77$      & $61.3\%$  & $1.26$        \\ \hline
CorrNet-256 (single)&C+H                              & $76.9\%$        & $0.82$      & $58.8\%$  & $1.31$        \\ \hline
CorrNet-256 &C+H                         & $78.9\%$        & $0.82$      & $64.8\%$  & $1.28$        \\ \hline
\end{tabular}}
\end{center}
\label{tab:abl_augmentations}
\end{table}

%------------------------------------------------------------------------------------------------------------------------------------------------------------------------------------------------------------------------
% ----- Conclusions and Future Work
%------------------------------------------------------------------------------------------------------------------------------------------------------------------------------------------------------------------------
\section{Conclusions}
%------------------------------------------------------------------------
%--- Conclusions
%------------------------------------------------------------------------
In this paper, we introduce CorrNet, a fully unsupervised approach for feature extraction. Our approach provides a flexible training strategy based on contrastive learning under spatial constraints, and a novel keypoint detection algorithm based on guided grad-CAM. Our experiments show CorrNet's robustness and advantages against previous methods under challenging illumination changes, achieving state-of-the-art results on HPatches~\cite{balntas2017hpatches}, and competitive results under viewpoint changes. We believe that contrastive learning will play a major role in many machine learning applications. The flexibility of setting up a training strategy that guides representation learning will potentially yield increasingly superior results on feature extraction in the wild. As a future direction, we will extend CorrNet to perform lifelong learning of visual features for continually improving its performance for pose estimation.

%------------------------------------------------------------------------------------------------------------------------------------------------------------------------------------------------------------------------
% ----- References
%------------------------------------------------------------------------------------------------------------------------------------------------------------------------------------------------------------------------
\bibliographystyle{IEEEtranS}
\bibliography{bib}
\end{document}